# Adaptative Bilingual Aligning Using Multilingual Sentence Embedding

**Olivier Kraif**

Univ. Grenoble Alpes, LIDILEM
F-38000 Grenoble, France
olivier.kraif@univ-grenoble-alpes.fr

**Abstract**

In this paper, we present an adaptive bitextual alignment system called Allign. This aligner relies on sentence embeddings to extract reliable anchor points that can guide the alignment path, even for texts whose parallelism is fragmentary and not strictly monotonic. In an experiment on several datasets, we show that Allign achieves results equivalent to the state of the art, with quasi-linear complexity. In addition, Allign is able to handle texts whose parallelism and monotonicity properties are only satisfied locally, unlike recent systems such as Vecalign or Bertalign.

## 1 Introduction

Sentence level bi-textual alignment consists in identifying sentences or groups of sentences that are translational equivalent between two texts. This translational equivalence can be considered from a broader point of view, and is not restricted to the case of a source text aligned with its translation into a target language: it can also concern two texts resulting from the translation of the same original text, in two different languages, or even in the same target language (this is why we speak here of bi-textual alignment rather than bilingual alignment).

Bi-textual alignment techniques appeared in the 1990s (Gale & Church, 1991; Brown et al., 1991) and subsequently played a major role in the development of Statistical Machine Translation (SMT), enabling the alignment of large parallel corpora. Although SMT has proved to be fairly robust to alignment errors in its parallel corpora, it has been shown that Neural Machine Translation (NMT) can suffer greatly from such errors (Khayrallah and Koehn, 2018). The quality of the bilingual alignment used as input to an NMT system therefore remains an issue, especially since, as Davis et al. (1993) had already pointed out, and as we have shown with the alignment of corpora from Wikipedia (Kraif, 2024, to appear), many translations are noisy and pose a challenge for sentence alignment.

In this paper, we propose an adaptive architecture based on two-stage alignment, using multilingual sentence embeddings to identify alignable areas before using more expensive dynamic programming methods.

In this way, we show that we can achieve the state of the art by considerably reducing the algorithmic cost, or even improve it for texts that do not respect monotonicity constraints.

## 2 Previous works

Historically, early systems relied on superficial cues such as sentence lengths, cognates (Church, 1993; Simard, Foster & Isabelle, 1992; McEnery & Oakes, 1996; Kraif, 2001; Lamraoui & Langlais, 2013), external bilingual lexicons (Varga et al., 2005) or lexicon derived from corpora (Moore, 2002).

More recently, several authors have shown that a new state of the art can be achieved using multilingual sentence embeddings, such as those of the LASER system (Artetxe and Schwenk, 2019) or LaBSE (Feng et al., 2020). Thomson & Koehn (2020), with the Vecalign system, develop a distance measure based on the cosine of compared sentence vectors, normalized by a random selection of vectors (they use LASER). They note that the cosine can be calculated for blocks of sentences, simply by summing the vectors. In this way, they propose to use a "recursive DP approximation" approach, to identify the best path step by step, first aligning blocks of sentences, then progressively increasing the resolution with smaller and smaller blocks. By applying this method recursively, they show that the complexity changes from quadratic to linear. In the evaluation of their system, they achieve the best results on the Text+Berg dataset (Volk et al., 2010) compared to 5 other competing aligners: Gale & Church (1991), BMA (Moore, 2002), Hunalign (Vargas, 2006), Bleualign (Sennrich & Volk, 2010), Gargantua (Braune & Fraser, 2010), and Coverage-Based (Gomes & Lopes, 2016). On a corpus extracted from the Bible dataset (Christodoupoulus & Steedman, 2015), they achieve a 28-point F1 improvement over Hunalign.

More recently, Liu & Zhu (2022), have proposed an architecture based on LaBSE vectors, with a two-stage strategy: first an optimal path is extracted based on 1-1, 0-1 and 1-0 matching; then n-n alignments are extracted between the points obtained, with a dynamic programing algorithm. On the Bible One dataset and an English-Chinese literary corpus, the Bertalign system obtained the best F1 score compared with 5 other systems, including Vecalign.

## 3 The Allign system

In order to reduce the search space, and to adapt to texts with significant parallelism breaks (deletions, additions, passage interchanges), we propose to implement the technique developed by Church (1993). Church proposed identifying reliable anchor points to guide the alignment path, based on point correspondences. In the absence of embeddings, the Char_align method relied on 4-gram character matches. Due to the high presence of cognates (named entities, dates, quantities) in the aligned sentence pairs, the alignment path is characterized by a high point density. A low-pass filter and thresholding are used to retain the best points around the path. Our idea is to apply the same method, but using a much richer and less noisy source of information than n-grams: multilingual embeddings.

We thus propose a two-stage architecture: first, we extract anchor points from sentence matches that have exceeded a certain similarity threshold; these anchor points enable us to identify alignable areas, when these anchor points are sufficiently dense and aligned along a local diagonal; then, inside alignable areas, we run a dynamic programming algorithm guided by these anchor points.

### 3.1 Extraction of anchor points

After calculating the embeddings of all sentences in both texts (with LaBSE, LASER or any encoder), a similarity matrix is calculated from the cosine of the vectors.

For each source sentence $S_i$, we calculate *k-Best($S_i$)* the *k* sentences *{$T_{j1}$ ..$T_{jk}$}* with the highest scores.

A margin criterion is then applied: the difference between the two best candidates must not be less than a threshold (*marginThreshold=0.05*), otherwise the candidates are ignored.

We then select only those candidates whose similarity exceeds a given threshold (*cosThreshold=0.4*).

We perform the same calculation for target sentences $T_j$, extract *k-Best($T_j$)* and apply the same criteria. Finally, we retain only those points *(i,j)* such that $i \in$ *k-Best($T_j$)* and $j \in$ *k-Best($S_i$)*.

For each candidate point, a high-pass filter is applied: the density of points in a zone centered on the point, parallel to the diagonal, with length deltaX (20) and height deltaY (3), is calculated. If the ratio between this density and the average density is below a certain threshold (minDensityRatio=0.3), the point is eliminated, as shown in figure 1.

A second filtering step then resolves conflicts when two or more points are located in the same row or column: only the point with the highest local density is retained.

### 3.2 Determining alignable intervals

Optionally, anchor points can be used to determine alignable intervals between source and target. We implemented the following algorithm:
- we denote *Anchors* the list of anchor points sorted by increasing horizontal coordinate
- for each anchor point *Anchors[i]=($x_i$,$y_i$)*, the deviation of *($x_i$,$y_i$)* from the diagonal passing through the previous point ($x_{i-1}$,$y_{i-1}$) is calculated as follow:

$$deviation = |(y_i - (y_{i-1} + (x_i - x_{i-1}) * sentRatio))|$$

The point *($x_i$,$y_i$)* is ignored in two cases:
- if it is not monotonic with respect to the two points preceding *Anchors[i-2], Anchors[i-1]* and two following *Anchors[i+1], Anchors[i+2]*, while these are monotonic ($x_{i-2} \leq x_{i-1} \leq x_{i+1} \leq x_{i+2}$ and $y_{i-2} \leq y_{i-1} \leq y_{i+1} \leq y_{i+2}$).

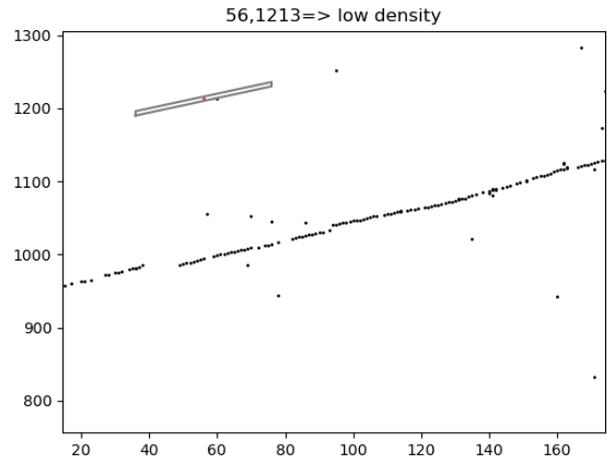

Figure 1: Elimination of candidate anchor points according to the local density

- if the deviation is greater than a threshold (set at 10) or $y_i < y_{i-1}$, and the local density ratio of *($x_i$,$y_i$)* is inferior to *minDensityRatio*.

If the deviation of *($x_i$,$y_i$)* is smaller than 20 (maxDistToTheDiagonal), it is included in the current interval.

Otherwise, a deviating point ($x_i$,$y_i$) can lead to the creation of a new interval in both cases:
- if *($x_i$,$y_i$)* is aligned with *($x_{i+1}$,$y_{i+1}$)* (in terms of deviation lower than the threshold) and its local density ratio is greater than *minDensityRatio*.
- if the Euclidean distance between *($x_{i-1}$,$y_{i-1}$)* and *($x_i$,$y_i$)* is greater than a certain threshold (*maxGapSize*, set at 100) and the density ratio is high (greater than 1.5**minDensityRatio)*.
A deviating point that does not satisfy one of these conditions is ignored.

When a new interval is created from *($x_i$,$y_i$)*, the previous one is closed until *($x_{i-1}$,$y_{i-1}$)*. Only intervals containing more than one anchor point and sufficiently dense (parameter *minHorizontalDensity* set at 0.15) are retained *in fine*.

Automatic interval extraction allows to implement an optional adaptive mode, useful when only part of the text is alignable: *charRatio* (ratio of sentence lengths, computed in characters) and *sentRatio* (ratio of number of sentences between target and source) are recalculated after the interval extraction step, and anchor point extraction is fully restarted with these new values (useful, in particular, for finely calculating the slope of local diagonals).

### 3.3 Dynamic programming step

Once the anchor points have been extracted (generally corresponding to a cloud of points fairly close together around the diagonal, as shown in Figure 1), and the alignment intervals defined, a dynamic programming algorithm is launched to calculate iteratively the optimal path leading to each anchor point.

Formally, a path is a sequence of parallel sentences groups. Permitted groupings are 1-0, 0-1, 1-1, 1-2, 2-1, ..., n-1, 1-n (*n* being controlled by the *maxGroupSize* parameter, set to 4 in our experiments). Optionally, we can also consider 2-2 groupings, and optionally ignore empty groupings (0-1, 1-0).

Calculating the best path means finding the best sequence of groups between the two ends of the alignable zone, in order to minimize a distance measure between each group.

This distance measure based on the sentence embeddings, for the source and target groups $g_i$ and $g_j$ is calculated as :

$$d_{embed}(g_i, g_j) = (1 - \cos(embed(g_i), embed(g_j)))$$

Unlike Bertalign, only individual phrases are encoded, the group vectors being obtained by simple addition of the sentence vectors. For empty groups (1-0), (0-1) we define a fixed distance of 1.

As in Liu & Zhu (2022), the cosine similarity measure is decreased according to the marginal cosines: we calculate the similarity of $g_i$ with the two sentences surrounding $g_j$, the similarity of $g_j$ with the two sentences surrounding $g_i$, then take the average of these similarities, and multiply it by a coefficient *c* (empirically set at 0.6).

$$d_{embed'}(g_i, g_j) = d_{embed}(g_i, g_j) + c * neighbourSim$$

$neighbourSim = 1/4 * (cos(embed(prec(g_i)), embed(g_j)) + cos(embed(succ(g_i)), embed(g_j)) + cos(embed(g_i), embed(prec(g_j))) + cos(embed(g_i), embed(succ(g_j))))$

where *prec(g_i)* is the sentence preceding $g_i$ and *succ(g_i)* is the sentence succeeding $g_i$.

Like Thomson & Koehn (2019), since cosine tends to favor large groups (e.g. 2-2 instead of twice 1-1) we apply a penalty proportional to the number of sentences involved.

$$d_{embed''}(g_i, g_j) = d_{embed'}(g_i, g_j) + p * (\text{size}(g_i) + \text{size}(g_j))$$

where the penalty factor *p* has been empirically set to 0.06.

To take sentence lengths into account (as in Gale & Church, 1991, but also Liu & Zhu, 2022), we define a second distance measure:

$$d_{length}(g_i, g_j) = 1 - np . \log_2(1 + \frac{len_{min}}{len_{max}})$$

where $len_{min}$= min(length($g_i$),length($g_j$)) and $len_{max}$= max(length($g_i$),length($g_j$)) and the character lengths are normalized using the *charRatio* parameter (which can be calculated automatically or set by the user).

The final distance measure is then calculated as the weighted sum of the two distances:

$$d(g_i, g_j) = (1-w) * d_{embed''}(g_i, g_j) + w * d_{length}(g_i, g_j)$$

Empirically, *w* has been set at 0.33.

Finally, each group's contribution to the total distance is multiplied by the number of sentences involved (so as not to favor long step paths over small step paths).

$$d_{final}(g_i, g_j) = d(g_i, g_j) * (\text{size}(g_i) + \text{size}(g_j))$$

In this way, the total distance of a path divided by the sum of the number of sentences in the two texts gives the average distance of each corresponding group, between 0 and 1: this score can be useful for indicating the relative similarity of texts after alignment.

To calculate the best path, we run a recursive calculation between each anchor point:
for each anchor point $(x_i, y_i)$ of *Anchors*, we calculate the deviation of $(x_i, y_i)$ from the diagonal passing through the previous point $(x_{i-1}, y_{i-1})$, and if this deviation is greater than a certain threshold (*localDiagBeam*) we ignore the point $(x_i, y_j)$. Otherwise, the recursive calculation of the best path leading to $(x_i, y_j)$ is launched (the optimal paths leading to $(x_{i-1}, y_{i-1})$ are stored in an array, so there's no need to recalculate them).

## 4 Experiment

As the Bertalign system represents the state of the art, we have only compared Allign results with this system.

In order to vary the language pairs and textual genres, we used the following datasets:

- Text+Berg (Volk et al., 2010): a corpus consisting of articles published in French and German by the Swiss Alpine Club and already used in several evaluation tasks.

- MD.fr-ar (Véronis et al., 2008): a corpus of *Le Monde diplomatique* articles translated from French into Arabic, and manually aligned for the Arcade 2 campaign.

- BAF (RALI, 1997): one of the first parallel English and French corpora to be manually aligned, including different text genres representing varying alignment difficulties.

- Grimm: the concatenation of the two volumes of Grimm fairy tales according to the 1857 edition of *Kinder und Hausmärchen*, and the *Contes choisis des frères Grimm* translated in 1864 by D. Baudry. This translation has the particularity of being a selection of 40 tales out of 210, reproduced in a completely different order to the original tales. For these two works, we produced a reference alignment at tale level (and not sentence level), in order to evaluate the detection of alignable intervals.

All the parameters where empirically fine tuned on a Chinese-French literary corpus, the novel

*Honggaoliang jiazun* from Mo Yan, that has been aligned manually[1]. On this development corpus, our system achieved a F-measure of 98.5 %.

We have then used the same settings for all the corpora, except for Grimm, for which we used the adaptive mode with interval detection. For this latter corpus, we will only give the results concerning the alignment of tales, given the lack of reference at sentence level.

## 5 Results and discussion

Precision, recall and F-measure values were calculated with strict scores from the script provided by Liu & Zhu (2022) on their repository[2].

|  | Bertalign | | | AIlign | | |
| --- | --- | --- | --- | --- | --- | --- |
| Dataset | P% | R% | F% | P% | R% | F% |
| Text+Berg | 93.2 | 94.1 | 93.6 | 91.3 | 93 | 92.1 |
| MD.ar-en | 95.0 | 95.7 | 95.4 | 94.6 | 96 | 95.3 |
| BAF | 92.1 | 95.6 | 93.8 | 92.4 | 94.5 | 93.4 |
| Grimm Tale level | | | | 98.7 | 92.8 | 95.7 |

Table 1: Comparative results of Bertalign and AIlign

|  | Bertalign | AIlign |
| --- | --- | --- |
| Text+Berg | 590 s. | 119 s. |
| MD.ar-en | 8114 s. | 2166 s. |
| BAF | 10882 s. | 1437 s. |

Table 2: Comparative execution time of Bertalign and AIlign[3]

### 5.1 Discussion

As a reminder, in the Arcade campaign (Véronis & Langlais, 2000), the best system obtained F=73.3% on the BAF corpus, with the literary and technical sub-corpuses posing serious difficulties. In the Arcade II campaign (Chiao et al., 2008), the best results for the MD.fr-ar sub-corpus were below 92%. For text+berg, Thomsonn & Koehn (2019) claim the best results for Vecalign among 9 evaluated systems, with F=90%. As we can see, AIlign's results are very close to Bertalign's ones, which represents the state-of-the-art system.

Its execution time is also significantly shorter than that of Bertalign. The most expensive part of the algorithm, for both, is the calculation of sentence embeddings by LaBSE. However, Ailign only calculates embeddings for individual sentences, and not for sequences of 1, 2, 3 or 4 consecutive sentences, which is a significant penalty on Bertalign's execution time. This difference can also explain the slightly better results of Bertalign: the encoding of group of sentences by LaBSE gives a more precise representation than the simple sum of individual vectors. But this slight gain (from 0.1 to 1.5 of F) has an important cost in performance (execution time is from 4 to 7 times longer).

Given the high density of anchor points, even between very distant languages (such as French and Arabic, or French and Chinese), the execution of the dynamic programming algorithm becomes quasi-linear. The only quadratic step is the calculation of the cosine for the similarity matrix, and the search for the *k-Best* points in row and column, but these calculations turn out to be very brief (less than one second) even for long texts, and generally represent only a very small fraction of the whole (e.g. 6 s. for Grimm). What's more, they can be easily parallelized on a GPU.

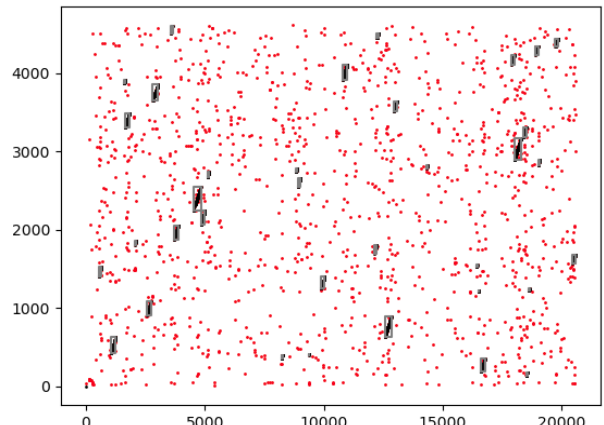

Figure 2: Identification of alignable intervals for the Grimm corpus (gray rectangles)

As far as the detection of alignable intervals is concerned, the results for the Grimm corpus are of very high quality (see Figure 2). Only three tales are completely missing: two very short ones (6 sentences for *Der undankbare Sohn*, 13 sentences for *Gottes Speise*) and one tale written in a German dialect (*Das Bürle im Himmel*). Half of a 4th dialectal tale (*Von dem Fischer un syner Fru*) is truncated, which explains a relative loss of recall, while precision remains very high.

## 6 Conclusion

In the field of bi-textual alignment, we have shown how to make the most of the new source of information provided by multilingual embeddings such as those produced by LaBSE. Thanks to these multilingual transformers, a new state of the art has been reached on this task, and it is now possible to obtain high quality alignment even for bi-texts with significant breaks in parallelism.

In future work, we plan to apply our approach to noisy translations from Wikipedia.

The AIlign codes and the datasets used in this article are freely available at: https://gricad-gitlab.univ-grenoble-alpes.fr/kraifo/

## 7 Bibliographical References

---

1 We would like to thank Yuhe Tang and Yinjie Wang who gave us the manually aligned version of Mo Yan's novel.
2 https://github.com/bfsujason/bertalign/tree/main
3 The script has been run on a laptop computer using Intel Core i7-6700HQ CPU @ 2.60GHz × 8, 16Go RAM.

## 8   Language Resource References